\documentclass[a4paper, 10 pt, conference]{ieeeconf}
\include{setup}
\usepackage{lipsum}  
 \usepackage{multirow}
 \usepackage{makecell}
 \usepackage{url}
\IEEEoverridecommandlockouts 
\newcommand{\vq}{\mathbf{q}}

\newcommand{\vy}{\mathbf{y}}
\newcommand{\vx}{\mathbf{x}}
\title{\LARGE \bf \myTitle
}

\author{\myName
	\thanks{All authors are with School of Informatics, University of Edinburgh (Informatics Forum, 10 Crichton Street, Edinburgh, EH8 9AB, United Kingdom). email: yiming.yang@ed.ac.uk}
}

\begin{document}

	\maketitle
	\thispagestyle{empty}
	\pagestyle{empty}

	\begin{abstract}
		Planning balanced and collision--free motion for humanoid robots is non--trivial, especially when they are operated in complex environments, such as reaching targets behind obstacles or through narrow passages. We propose a method that allows us to apply existing sampling--based algorithms to plan trajectories for humanoids by utilizing a customized state space representation, biased sampling strategies, and a steering function based on a robust inverse kinematics solver. Our approach requires no prior offline computation, thus one can easily transfer the work to new robot platforms. We tested the proposed method solving practical reaching tasks on a 38 degrees--of--freedom humanoid robot, NASA Valkyrie, showing that our method is able to generate valid motion plans that can be executed on advanced full--size humanoid robots. We also present a benchmark between different motion planning algorithms evaluated on a variety of reaching motion problems. This allows us to find suitable algorithms for solving humanoid motion planning problems, and to identify the limitations of these algorithms.
	\end{abstract}
	
	\IEEEpeerreviewmaketitle

\section{Introduction}
\label{sec:intro}
Humanoid robots are highly redundant systems that are designed for accomplishing a variety of  tasks in environments designed for human. However, humanoids have a large number of degrees--of--freedom (DoF) which makes motion planning extremely challenging. In general, optimization--based algorithms are suitable for searching for optimal solutions even in high dimensional systems \cite{aico} \cite{chomp}, but it is non--trivial to generate optimal collision--free trajectories for humanoids using optimization approaches, especially in complex environments. This is mainly due to the highly non--linear map between the robot and the collision environment. This mapping can be modelled in some abstract spaces to provide real--time collision avoidance capabilities on low DoF robotic arms \cite{dmesh} difficult for high DoF humanoids due to the curse of dimensionality and it often causes local minima problems. Additionally, solving locomotion and whole--body manipulation in complex environments as one combined problem requires searching through a large space of possible actions. Instead, it is more effective to first generate robust walking plans to move the robot to a desired standing location, and then generate collision--free motion with stationary feet \cite{idrm}. Although assuming fixed feet position may be viewed as restrictive, we argue that a large variety of whole body manipulation tasks can still be executed as a series of locomotion and manipulation subtasks. We propose an extension to a family of sampling based motion planning algorithms that will allow us to plan collision--free whole--body motions on floating based systems which require active balancing.

\begin{figure}[t]
	\centering
	\includegraphics[width=\linewidth]{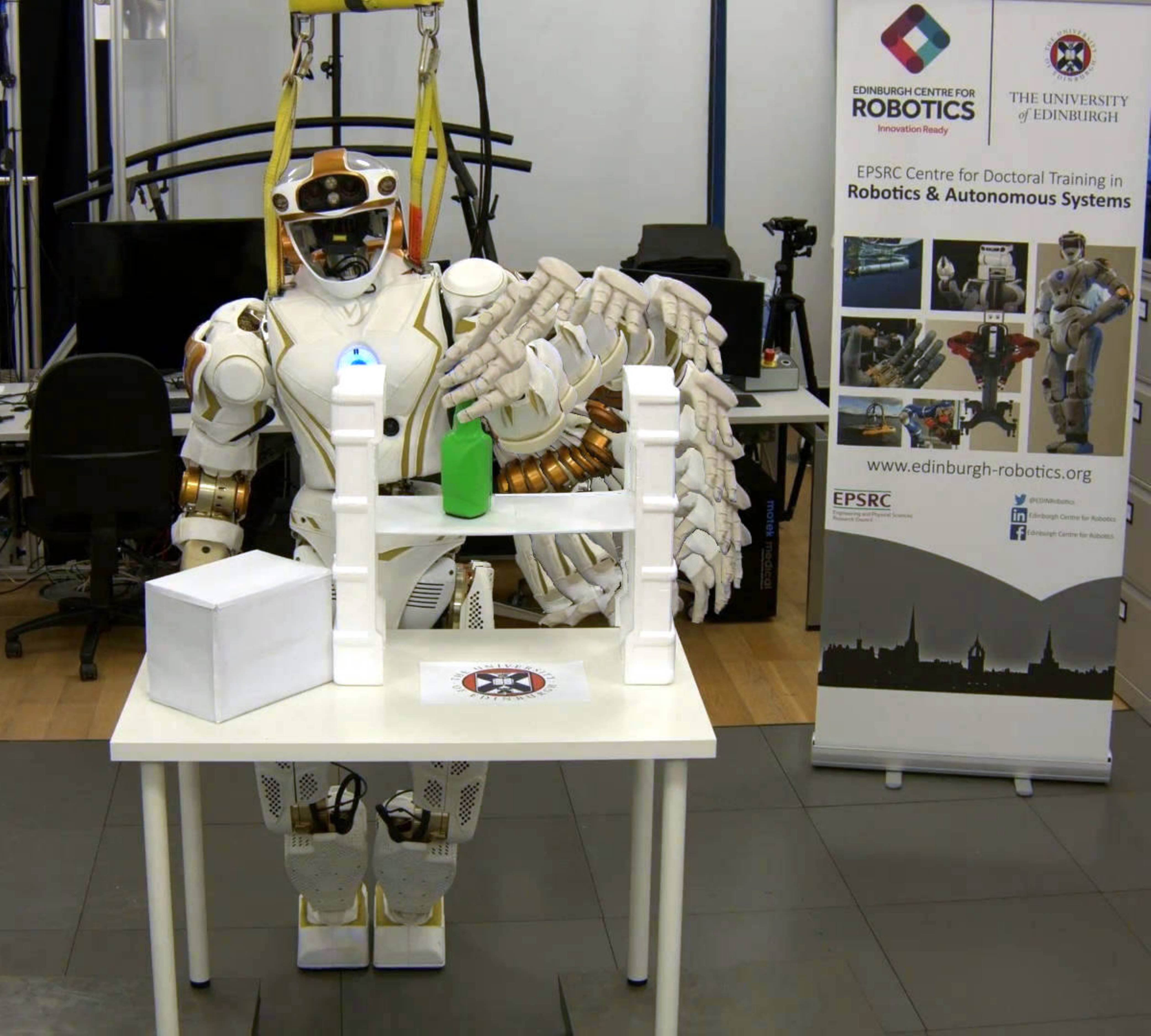}
	\caption{Collision--free and balanced whole--body motion executed on the 38 DoF NASA Valkyrie robot.}
	\label{fig:intro}
\end{figure}

Sampling--based planning (SBP) algorithms, such as RRT \cite{rrt1998} and PRM \cite{prm1996}, are capable of efficiently generating globally valid collision-free trajectories due to their simplicity. In the past two decades, SBP algorithms have been applied to countless problems with a variety of derivatives, such as RRT-Connect \cite{rrtconnect}, Expansive Space Trees (EST, \cite{est1997}), RRT*/PRM* \cite{rrtstar}, Kinematic Planning by Interior-Exterior Cell Exploration (KPIECE, \cite{csucan2009kinodynamic}), and many others \cite{6722915}. However, since the SBP algorithms were originally designed for mobile robots and low DoF robotic arms, using them on high DoF systems requiring active balancing is still challenging. We will call a robot pose statically balanced if the controller can achieve an equilibrium in this state while achieving zero velocity and acceleration (e.g. when the projection of centre of mass lies within the support polygon). The subset of robot configurations with this property forms a low dimensional manifold defined by the balance constraint. In practice, the rejection rate of random samples is prohibitively high without the explicit or implicit knowledge of the manifold. Approaches have been proposed to address this particular problem of using SBP algorithms for humanoid robots. Kuffner at al. \cite{kuffner2005motion} use a heavily customized RRT--Connect algorithm to plan whole body motion for humanoids, where they only sample from a pre-calculated pool of postures for which the robot is in balance. Hauser at al. \cite{hauser2008using} introduce motion primitives into SBP algorithms where the sampler only samples states around a set of pre--stored motion primitives. A similar approach is used in \cite{7363504} with centre--of--mass (CoM) movement primitives. These approaches share the common idea of using an offline generated sample set to bootstrap online processes, thus allowing algorithms to bypass the expensive online generation of balanced samples. However, this leads to the problem where one has to store a significant number of samples to densely cover the balance manifold, otherwise the algorithms may fail while valid solutions exist but were not stored during offline processing. Another issue is that the pre--processing is normally platform specific, which makes it difficult and time consuming to transfer the work to other robot platforms.


To this end, instead of developing new SBP algorithms specifically for humanoids, we focus on enabling the standard SBP algorithms to solve humanoids motion planning problems by modifying the underlying key components of generic SBP approaches, such as \emph{space representation}, \emph{sampling strategies} and \emph{interpolation functions}. In order to make the method generic for any humanoid platforms, rather than store balanced samples during offline processing, we use a non--linear optimization based \cite{snopt} whole--body inverse kinematics (IK) solver to generate balanced samples on--the--fly. Thus, the proposed method can be easily applied to different humanoid robot platforms without extensive pre--processing and setup. We evaluate the proposed method on a 36 DoF Boston Dynamics Atlas and a 38 DoF NASA Valkyrie humanoid robots, to show that our method is capable of generating reliable collision--free whole--body motion for a generic humanoid. We also evaluate the difference between sampling in end--effector and configuration spaces for different scenarios, and compare the planning time and trajectory length to find an optimal trade off between efficiency and optimality. In particular, we apply our work to solve practical reaching tasks on the Valkyrie robot, as highlighted in Fig.~\ref{fig:intro}, showing that the proposed method can generate reliable whole--body motion that can be executed on full--size humanoid robots.

\section{Problem Formulation}
Let $\mathcal{C} \in \mathbb{R}^{N+6}$ be a robot's configuration space with $N$ the number of articulated joints and the additional 6-DoF of the under actuated virtual joint that connects the robot's pelvis ($T_\mathit{pelvis}$) and the world $\mathcal{W}\in SE(3)$. Let $\vq\in\mathcal{C}$ be the robot configuration state, $\mathcal{C}_\mathit{balance}\subset\mathcal{C}$ the manifold that contains statically balanced configurations, $\mathcal{C}_\mathit{free}\subset\mathcal{C}$ the manifold contains collision free configurations and $\mathcal{C}_\mathit{valid}\equiv\mathcal{C}_\mathit{balance}\cap\mathcal{C}_\mathit{free}$ the valid configuration manifold. 

For humanoid robots, valid trajectories can only contain states from valid configuration manifold, i.e. $\vq_{[0:T]}\subset\mathcal{C}_\mathit{valid}$. Generating collision free samples is straightforward by using random sample generators and standard collision checking libraries. However, generating balanced samples is non-trivial, where a random sampling technique is incapable of efficiently finding balanced samples on the low dimensional manifold $\mathcal{C}_\mathit{balance}$ by sampling in high dimensional configuration space $\mathcal{C}$. Guided sampling or pre--sampling process is required for efficient valid sample generation. In our approach, a whole-body inverse kinematic solver is employed to produce statically balanced samples. Static balance constraint is a combination of feet and CoM poses constraints, i.e. the static balance constraint is considered as satisfied when the robot's feet has stable contacts with ground and the CoM ground projection stays within the support polygon. 

\subsection{Whole--body Inverse Kinematics}
Given a seed configuration $\vq_\mathit{seed}$ and nominal configuration $\vq_\mathit{nominal}$ and a set of constraints $\mathbf{C}$, an output configuration that satisfies all the constraints can be generally formulated as:
\begin{equation}
	\vq^* = \mathit{IK}(\vq_\mathit{seed}, \vq_\mathit{nominal},\mathbf{C})
	\label{eq:ik}
\end{equation}
The Constraint set for a whole--body humanoid robot may include single joint constraints, such as position and velocity limits for articulated joints, it may also include workspace pose constraints, e.g. end--effector poses, centre-of-mass position. In the rest of the paper, unless specified otherwise, we assume the quasi--static balance constraint and joint limits constraints are included in $\mathbf{C}$ by default. We formulate the IK problem as a non-linear optimization problem (NLP) of form:
\begin{eqnarray}
	\vq^*  = & \arg\min_{\vq\in\mathbb{R}^{N+6}} \|\vq-\vq_\mathit{nominal}\|^2_{Q_q}\\
	\textnormal{subject to} & \mathbf{b}_l\leq \vq	\leq \mathbf{b}_u\\
	& c_i(\vq)\leq 0, c_i\in \mathbf{C}
	\label{eq:iksqp}
\end{eqnarray}
where $Q_q\succeq 0$ is the weighting matrix, $\mathbf{b}_l$ and $\mathbf{b}_u$ are the lower and upper joint bounds. We use a randomly sampled state as the seed pose $\vq_\mathit{seed}$. We then use this pose as the initial value in the first iteration of SQP solver. Depending on the implementation of the SBP algorithm, we either choose $\vq_\mathit{nominal}$ to be the current robot state or one of the neighbouring poses drawn from the pool of candidate poses already explored by the SBP algorithm.

\subsection{Sampling--based Motion Planning}
Let $\vx\in \mathcal{X}$ be the space where the sampling is carried out. The planning problem can be formulated as 
\begin{equation}
	\vq_{[0:T]} = Planning(\mathbf{Rob}, \mathbf{Env}, \vx_0,\vx_T)
	\label{eq:planning}
\end{equation}
where $\mathbf{Rob}$ is the robot model and $\mathbf{Env}$ is the environment instance in which this planning problem is defined. $\vx_0$ and $\vx_T$ are the initial and desired states. 

\begin{figure}[t]
	\centering
	\includegraphics[width=0.48\textwidth]{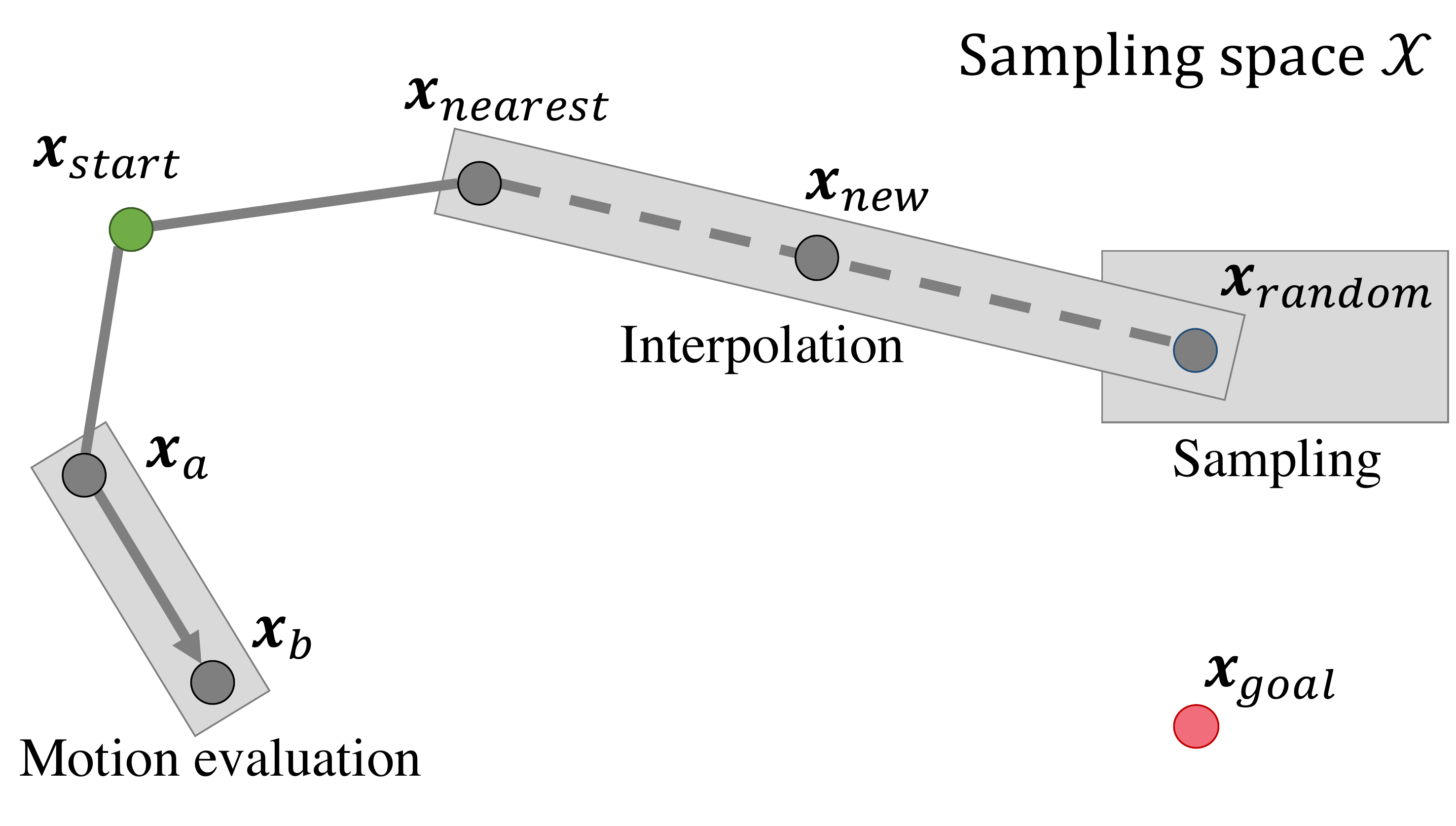}
	\caption{Instead of developing new algorithms, we modify those underlying components in SBP solvers to make standard algorithms be capable of solving motion planning problems for humanoid robots.}
	\label{fig:comp}
\end{figure}

In order for SBP algorithms to be able to plan motions for humanoid robots, we need to modify the following components that are involved in most algorithms as shown in Fig.~\ref{fig:comp}: the space $\mathcal{X}$ where the sampling is carried out; the strategies to draw random samples; and the interpolation function which is normally used in steering and motion evaluation steps. In the next section, we will discuss the details of modifications we applied on those components for scaling standard SBP algorithms to humanoids.

\section{Sampling--based Planning for Humanoids}
We separate the work into two parts, configuration space sampling and end--effector space sampling. In configuration space sampling approach, the state is represented in $\mathbb{R}^{N+6}$ space with joint limits and maximum allowed base movement as the bounds, the sampling state is identical to robot configuration, i.e. $\vx=\vq\in\mathcal{C}$. For reaching and grasping problems, one might be interested in biasing the sampling in the end--effector related constraints, e.g. to encourage shorter end--effector traverse distance. The end--effector space approach samples in $SE(3)$ space with a region of interests around the robot as the bounds, the state is equivalent to the end--effector's forward kinematics, i.e. $\vx=\Phi(\vq)\in\mathcal{W}$ where $\Phi(\cdot)$ is the forward kinematics mapping. However, the final trajectories are represented in configuration space, thus we associate a corresponding configuration for each end--effector space state to avoid ambiguity and duplicated calls of IK solver.
\subsection{Configuration Space Sampling}
Algorithm~\ref{alg:cspace} highlights the components' modifications required for sampling in configuration space:

\begin{algorithm}[t]
	\caption{Humanoid Configuration Space SBP}
	\label{alg:cspace}
	\textbf{sampleUniform}()
	\begin{algorithmic}[1]
		\State $\mathit{succeed}=\mathrm{False}$
		
		\While{\textbf{not} $\mathit{succeed}$}
			\State $\bar\vq_\mathit{rand}=\mathit{RandomConfiguration}()$
			\State	$\vq_\mathit{rand},\mathit{succeed} = \mathit{IK}(\bar\vq_\mathit{rand}, \bar\vq_\mathit{rand},\mathbf{C})$
		\EndWhile
		
		\noindent \Return {$\vq_\mathit{rand}$}
	\end{algorithmic}
	\hrulefill
	
	\textbf{sampleUniformNear}($\vq_\mathit{near},d$)
	\begin{algorithmic}[1]
		\State $\mathit{succeed}=\mathrm{False}$
		
		\While{\textbf{not} $\mathit{succeed}$}
			\State $A\leftarrow \mathit{Zeros}(N+6)$
			\While{\textbf{not} $\mathit{succeed}$}
				\State $\bar\vq_\mathit{rand}=\mathit{RandomNear}(\vq_\mathit{near},d)$
				\State Set constraint $\Vert \vq_\mathit{rand}-\bar\vq_\mathit{rand}\Vert_W < A$
				\State	$\vq_\mathit{rand},\mathit{succeed} = \mathit{IK}(\bar\vq_\mathit{rand}, \vq_\mathit{near},\mathbf{C})$
				\State Increase $A$
			\EndWhile
				\If {$\mathit{distance}(\vq_\mathit{rand},\vq_\mathit{near})>d$}
					\State $\mathit{succeed}=\mathrm{False}$
				\EndIf
		\EndWhile
		
		\noindent \Return {$\vq_\mathit{rand}$}
	\end{algorithmic}
	\hrulefill
	
	\textbf{interpolate}($\vq_a,\vq_b, d$)
	\begin{algorithmic}[1]
		\State $\bar\vq_\mathit{int}=\mathit{InterpolateConfigurationSpace}(\vq_a,\vq_b,d)$
		\State $\mathit{succeed}=\mathrm{False}$
		\State $A\leftarrow \mathit{Zeros}(N+6)$
		\While{\textbf{not} $\mathit{succeed}$}
			\State Set constraint $\Vert \vq_\mathit{int}-\bar\vq_\mathit{int}\Vert_W < A$
			\State	$\vq_\mathit{int},\mathit{succeed} = \mathit{IK}(\bar\vq_\mathit{int}, \vq_a,\mathbf{C})$
			\State Increase $A$
		\EndWhile
		
		\noindent \Return {$\vq_\mathit{int}$}
	\end{algorithmic}
\end{algorithm}

\subsubsection{Sampling Strategies}
For $\mathit{sampleUniform}()$, we first generate random samples from $\mathcal{X}$ and then use fullbody IK solver to process the random samples to generate samples from the balanced manifold $\mathcal{X}_\mathit{balance}$
\begin{equation}
	\vq_\mathit{rand} = \mathit{IK}(\bar\vq_\mathit{rand}, \bar\vq_\mathit{rand},\mathbf{C})
\end{equation}
where $\bar\vq_\mathit{rand}\in\mathcal{X}$ is a uniform random configuration and $\vq_\mathit{rand}\in\mathcal{X}_\mathit{balance}$ is random sample from the balanced manifold. We use $\bar\vq_\mathit{rand}$ as nominal pose since we want to generate random postures rather than postures close to other already existing samples. This is to indirectly encourage exploration of the null-space of the task. The constraint set $\mathbf{C}$ contains static balance constraint and joint limits constraints. When sampling around a given state, $\mathit{sampleUniformNear}(\vq_\mathit{near},d)$, we first get a random state $\bar\vq_\mathit{rand}$ that is close to $\vq_\mathit{near}$ within distance $d$. The IK solver is invoked with $\bar\vq_\mathit{rand}$ as the seed pose, and $\vq_\mathit{near}$ as the nominal pose. An additional configuration space constraint is added to the constraint set
\begin{equation}
	\|\vq_\mathit{rand}-\bar\vq_\mathit{rand}\|_W \leq A
	\label{eq:poscon}
\end{equation}
where $A\in\mathbb{R}^{N+6}$ is a tolerance vector initially set to zero. In most cases the system will be over constrained, in which case we need to increase the tolerance to ensure balance. Normally, the lower--body joints are neglected first, i.e. increasing corresponding $w_i$, meaning that we allow the lower--body joints to deviate from $\bar\vq_\mathit{rand}$ in order to keep feet on the ground and maintain balance. We use $\vx_\mathit{near}$ as the nominal pose since later on the random state is likely to be appended to $\vq_\mathit{near}$, where one wants the random state be close to the near state. The new sample is discarded if the distance between $\vq_\mathit{near}$ and $\vq_\mathit{rand}$ exceeds the limit $d$.

\subsubsection{Interpolation}
In order to find a balanced state interpolated along two balanced end--point states, we first find the interpolated, likely to be un--balanced state 
\begin{equation}
	\bar\vq_\mathit{int} = \vq_a + d\|\vq_b-\vq_a\|.
\end{equation}
A similar configuration space constraint to (\ref{eq:poscon}) is applied to constrain the balanced interpolated state $\vq_\mathit{int}$ close to $\bar\vq_\mathit{int}$
\begin{equation}
	|\vq_\mathit{int}-\bar\vq_\mathit{int}\| \leq A
\end{equation}
The two end-point states $\vq_a$ and $\vq_b$ are valid samples generated using our sampling strategies. Due to the convex formulation of the balance constraint, a valid interpolated state is guaranteed to be found. It is worth mentioning that in some cases the interpolation distance equation no longer holds after increasing the tolerance, i.e. $\frac{\|\vx_\mathit{int}-\vx_a\|}{\|\vx_b-\vx_a\|}\neq d$. However, this is a necessary step to ensure that the balance constraint are satisfied. 

\subsection{End-Effector Space Sampling}
\begin{algorithm}[t]
	\caption{Humanoid End--Effector Space SBP}
	\label{alg:espace}
	\textbf{sampleUniform}()
	\begin{algorithmic}[1]
		\State $\mathit{succeed}=\mathrm{False}$
		
		\While{\textbf{not} $\mathit{succeed}$}
			\State $\bar\vx_\mathit{rand}=\mathit{RandomSE3}()$
			\State Set constraint $\Vert\bar\vx_\mathit{rand}-\Phi(\vq_\mathit{rand})\Vert \leq 0$
			\State	$\vq_\mathit{rand},\mathit{succeed} = \mathit{IK}(\bar\vq_\mathit{rand}, \bar\vq_\mathit{rand},\mathbf{C})$
		\EndWhile
		\State $\vx_\mathit{rand} = \Phi(\vq_\mathit{rand})$
		
		\noindent \Return {$\vx_\mathit{rand}, \vq_\mathit{rand}$}
	\end{algorithmic}
	\hrulefill
	
	\textbf{sampleUniformNear}($\vx_\mathit{near},d$)
	\begin{algorithmic}[1]
		\State $\mathit{succeed}=\mathrm{False}$
		
		\While{\textbf{not} $\mathit{succeed}$}
			\State $\bar\vx_\mathit{rand}=\mathit{RandomNearSE3}(\vx_\mathit{near},d)$
			\State Set constraint $\Vert\bar\vx_\mathit{rand}-\Phi(\vq_\mathit{rand})\Vert \leq 0$
			\State	$\vq_\mathit{rand},\mathit{succeed} = \mathit{IK}(\vq_\mathit{rand}, \vq_\mathit{near},\mathbf{C})$
		\EndWhile
		\State $\vx_\mathit{rand}=\bar\vx_\mathit{rand}$
		
		\noindent \Return {$\vx_\mathit{rand},\vq_\mathit{rand}$}
	\end{algorithmic}
	\hrulefill
	
	\textbf{interpolate}($\vx_a,\vx_b, d$)
	\begin{algorithmic}[1]
		\State $\bar\vx_\mathit{int}=\mathit{InterpolateSE3}(\vx_a,\vx_b,d)$
		\State $\mathit{succeed}=\mathrm{False}$
		\State $B\leftarrow \mathit{Zeros}(SE3)$
		\While{\textbf{not} $\mathit{succeed}$}
			\State Set constraint $\Vert\bar\vx_\mathit{int}-\Phi(\vq_\mathit{int})\Vert < B$
			\State	$\vq_\mathit{int},\mathit{succeed} = \mathit{IK}(\vq_a, \vq_a,\mathbf{C})$
			\State Increase $B$
		\EndWhile
		\State $\vx_\mathit{int}=\Phi(\vq_\mathit{int})$
		
		\noindent \Return {$\vx_\mathit{int},\vq_\mathit{int}$}
	\end{algorithmic}
\end{algorithm}
Algorithm~\ref{alg:espace} highlights the components' modifications required for sampling in end--effector space:
\subsubsection{Sampling Strategies}
It is straight forward to sample in $SE(3)$ space, however, it is non--trivial to sample balanced samples from the $\mathcal{X}_\mathit{balance}$ manifold. For $\mathit{sampleUniform()}$, we first randomly generate $SE(3)$ state $\bar\vx_\mathit{rand}$ within a region of interest in front of the robot. The whole--body IK is invoked with an additional end--effector pose constraint
\begin{equation}
	\Vert\bar\vx_\mathit{rand}-\Phi(\vq_\mathit{rand})\Vert \leq 0
	\label{eq:effik}
\end{equation}
The sampler keeps drawing new random states $\bar\vx_\mathit{rand}$ until the SQP solver returns a valid output $\vq^*$. The valid random state $\vx_\mathit{rand}$ can be calculated using forward kinematics, e.g. $\vx_\mathit{rand} = \Phi\left(\vq^*\right)$. The same procedure applies to $\mathit{sampleNear}(\vx_\mathit{near},d)$, but using $\vx_\mathit{near}$ as the seed configuration.

\subsubsection{Interpolation}
Similar to sampling near a given state, for interpolation in end--effector space, we first find the interpolated state $\bar\vx_\mathit{int}\in SE(3)$ and add the following term into constraint set
\begin{equation}
	\|\bar\vx_\mathit{int}-\Phi(\vq)\| \leq B
\end{equation}
where $B\in \mathbb{R}^6$ is a tolerance vector initially set to zero. If the system is over constrained after adding end--effector pose constraint, we need to selectively relax the tolerance for different dimensions ($x,y,z,roll,pitch,yaw$) until the IK solver succeeds. Then we reassign the interpolated state using forward kinematics, $\vx_\mathit{int} = \Phi(\vq_\mathit{int})$.

\subsubsection{Multi-Endeffector Motion Planning}
Some tasks require coordinated motion involving multiple end--effectors, e.g. bi--manual manipulation and multi--contact motion. It is obvious that, from a configuration space point of view, there is no difference as long as the desired configuration is specified. It is also possible for end--effector space sampling approach to plan motion with multiple end--effector constraints. Let $\vy_k^*\in SE(3)$ be the desired pose constraints for end--effector $k\in\{1,\dots,K\}$. A meta end--effector space $\mathcal{X}\in\mathbb{R}^{6\times K}$ can be constructed to represent the sampling space for all end--effectors. Similar sampling and interpolation functions can be implemented by constructing extra constraints for each end--effector $k$.

\begin{table}[t!]
	\centering
	\caption{Planning time of empty space reaching problem crossing different algorithms, in seconds.}
	\label{table:freereach}
	\begin{tabular}{|p{1.65cm}|p{2.86cm}|p{2.88cm}|}
		\hline
		\multirow{2}{*}{Algorithms} & \multicolumn{2}{c|}{Sampling Space}        \\ \cline{2-3}
		& End--Effector Space & Configuration Space \\ \hline\hline
		RRT                         & $25.863\pm 22.894$  & $1.4129\pm 1.4466$    \\ 
		PRM                        & $4.2606\pm 3.0322$ 			& $0.5912\pm 0.5912$    \\
		EST                         &  $28.055\pm 18.270$                  & $0.3112\pm 0.3112$                    \\ \hline\hline
		BKPIECE               &  $5.3989\pm 5.9470$                   & $0.1781\pm 0.0332$                    \\
		SBL                         &  $3.0602\pm 0.9859$                   & $0.2804\pm 0.0480$                    \\
		RRT--Connect          &  $2.8228\pm 0.3412$                   & $0.1853\pm 0.0450$                   \\
		\hline
	\end{tabular}
\end{table}

\section{Evaluation}
We aim to generalize the common components of sampling--based motion planning algorithms for humanoid robots so that existing algorithms can be used without extra modification. We implemented our approach in the EXOTica motion planning and optimization framework \cite{exotica} as humanoid motion planning solver, which internally invokes the SBP planners from the Open Motion Planning Library (OMPL, \cite{ompl}). We have set up the system with our customized components, and evaluated our approach on the following 6 representative algorithms: RRT \cite{rrt1998}, RRT-Connect \cite{rrtconnect}, PRM \cite{prm1996}, BKPIECE \cite{csucan2009kinodynamic}, EST \cite{est1997}) and SBL \cite{sbl}. The evaluations are performed on a single thread of the 4.0 GHz Intel Core i7-6700K CPU.

\newcolumntype{?}{!{\vrule width 1pt}}
\begin{table*}[t!]
	\centering
	\caption{Evaluation of whole--body collision--free motion planning. RRT--Connect$_e$ sampling in end--effector space, all other methods sampling in configuration space. $\mathcal{C}$ cost is the configuration space trajectory length, $\mathcal{W}$ cost is the end--effector traverse distance in workspace, CoM cost is the CoM traverse distance in workspace. No. evaluation shows the number of state evaluation calls, i.e. evaluate if a sampled/interpolated state is valid. No. IK indicates the number of online whole--body IK calls, and IK time is the total time required for solving those IK, which is the most time consuming element. The result is averaged over 100 trails.}
	\label{table:results}
	\begin{tabular}{?c|l?c?c|c|c|l|l|l?}
		\Xhline{3\arrayrulewidth} 
		Tasks                   & Algorithms          & Planning time (s)  & $\mathcal{C}$ cost (rad.)  & $\mathcal{W}$ cost (m) & CoM cost (m) & No. evaluation & No. IK  & IK time (s)\\ \Xhline{3\arrayrulewidth} 
		\multirow{4}{*}{Task 1} & BKPIECE$_c$	      & 42.5 $\pm$ 26.4 & 7.37 $\pm$ 2.43 & 2.10 $\pm$ 0.80 & 0.24 $\pm$ 0.10 & 1946 $\pm$ 1207 & 2598 $\pm$ 1582 & 41.4 $\pm$ 25.7\\ \cline{2-9} 
		& SBL$_c$	          & 27.8 $\pm$ 8.59& 6.25 $\pm$ 1.06 & 2.14 $\pm$ 0.71 & 0.23 $\pm$ 0.06 & 1313 $\pm$ 418 & 1508 $\pm$ 445 & 27.0 $\pm$ 8.33\\ \cline{2-9} 
		& RRT--Connect$_e$	  & 9.91 $\pm$ 4.80& 2.93 $\pm$ 0.96 & \textbf{0.58} $\pm$ 0.11 & \textbf{0.07} $\pm$ 0.02 & 597 $\pm$ 354 & 727 $\pm$ 387 & 9.51 $\pm$ 4.58\\ \cline{2-9} 
		& RRT--Connect$_c$    & \textbf{1.53} $\pm$ 0.80 & \textbf{2.71} $\pm$ 0.68 & 0.99 $\pm$ 0.23 & 0.11 $\pm$ 0.03 & \textbf{95} $\pm$ 54 & \textbf{118} $\pm$ 64 & \textbf{1.48} $\pm$ 0.77\\ \Xhline{3\arrayrulewidth} 
		\multirow{4}{*}{Task 2} & BKPIECE$_c$         & 40.5  $\pm$ 21.7 & 6.59 $\pm$ 2.43 & 1.95 $\pm$ 0.59 & 0.27 $\pm$ 0.09 & 1911 $\pm$ 970 & 2473 $\pm$ 1254 & 39.4 $\pm$ 20.1\\ \cline{2-9} 
		& SBL$_c$             & 22.2 $\pm$ 9.51 & 5.34 $\pm$ 2.00 & 1.79 $\pm$ 0.80 & 0.24 $\pm$ 0.09 & 1089 $\pm$ 472 & 1259 $\pm$ 547 & 21.5 $\pm$ 9.23\\ \cline{2-9} 
		& RRT--Connect$_e$    & 12.4 $\pm$ 6.65 & 4.12 $\pm$ 2.02 & \textbf{0.77} $\pm$ 0.08 & \textbf{0.09} $\pm$ 0.04 & 656 $\pm$ 405 & 826 $\pm$ 458 & 11.9 $\pm$ 6.41\\ \cline{2-9} 
		& RRT--Connect$_c$    & \textbf{2.25} $\pm$ 0.85 & \textbf{3.29} $\pm$ 1.14 & 1.20 $\pm$ 0.33 & 0.14 $\pm$ 0.05 & \textbf{106} $\pm$ 42 & \textbf{166} $\pm$ 59& \textbf{2.19} $\pm$ 0.83 \\ \Xhline{3\arrayrulewidth} 
		\multirow{4}{*}{Task 3} & BKPIECE$_c$	      & 45.7 $\pm$ 19.8 & 7.49 $\pm$ 2.52 & 1.96 $\pm$ 0.73 & 0.25 $\pm$ 0.08 & 2057 $\pm$ 949 & 2758 $\pm$ 1166 & 44.5 $\pm$ 19.3\\ \cline{2-9} 
		& SBL$_c$             & 33.8 $\pm$ 22.2& 8.68 $\pm$ 2.26 & 2.10 $\pm$ 0.44 & 0.28 $\pm$ 0.11 & 1414 $\pm$ 950 & 1756 $\pm$ 1151 & 33.0 $\pm$ 21.6\\ \cline{2-9} 
		& RRT--Connect$_e$    & 25.3 $\pm$ 13.9 & 7.19 $\pm$ 4.93 & \textbf{0.92} $\pm$ 0.13 & 0.16 $\pm$ 0.05 & 1031 $\pm$ 532 & 1436 $\pm$ 720 & 24.6 $\pm$ 13.7\\ \cline{2-9} 
		& RRT--Connect$_c$    & \textbf{3.45} $\pm$ 0.77 & \textbf{4.68} $\pm$ 0.59 & 1.38 $\pm$ 0.12 & \textbf{0.14} $\pm$ 0.03 & \textbf{165} $\pm$ 49 & \textbf{200} $\pm$ 53 & \textbf{3.36} $\pm$ 0.75\\ \Xhline{3\arrayrulewidth} 
	\end{tabular}
\end{table*}

\subsection{Empty Space Reaching}
In the first experiment, we have the robot reach a target pose in front of the robot in free space, where only self--collision and balance constraints are considered. This is a sanity check to show that the proposed method can be used robustly across different planning algorithms to generate trajectories for humanoid robots. We solve the reaching problem using the 6 testing algorithms in two different sampling spaces, each across 100 trials. The results are shown in Table~\ref{table:freereach}. Although the planning time varies across different algorithms and sampling spaces, the result shows that standard planning algorithms are able to generate motion plans for humanoid robots using our method. However, as expected, bi--directional algorithms are more efficient than their unidirectional variants. Also, sampling in configuration space is much more efficient than in end--effector space due to the higher number of IK calls.

\subsection{Collision--free Reaching}
\begin{figure}[t]
	\centering
	\includegraphics[width=0.15\textwidth]{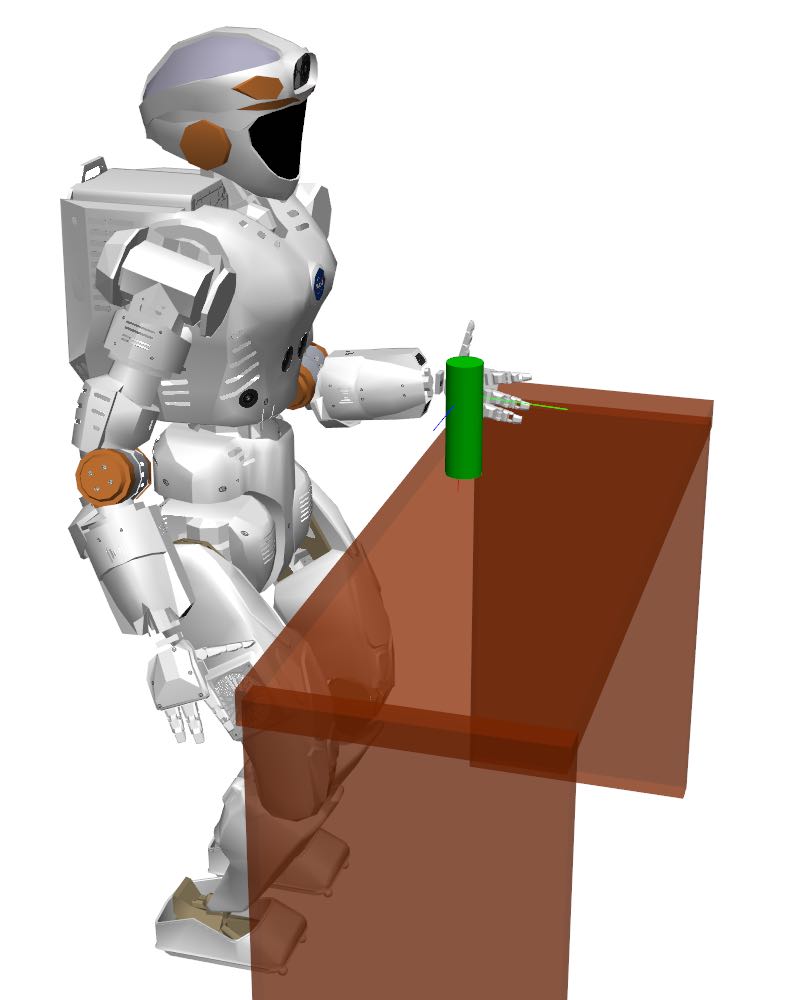}
	\includegraphics[width=0.15\textwidth]{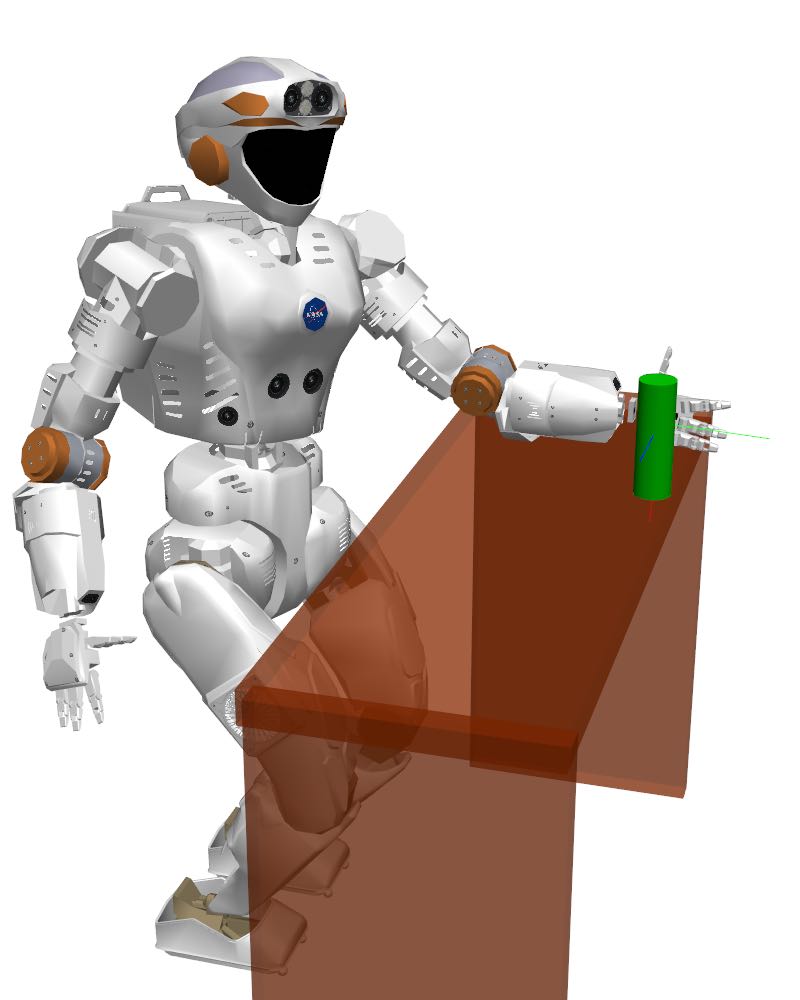}
	\includegraphics[width=0.15\textwidth]{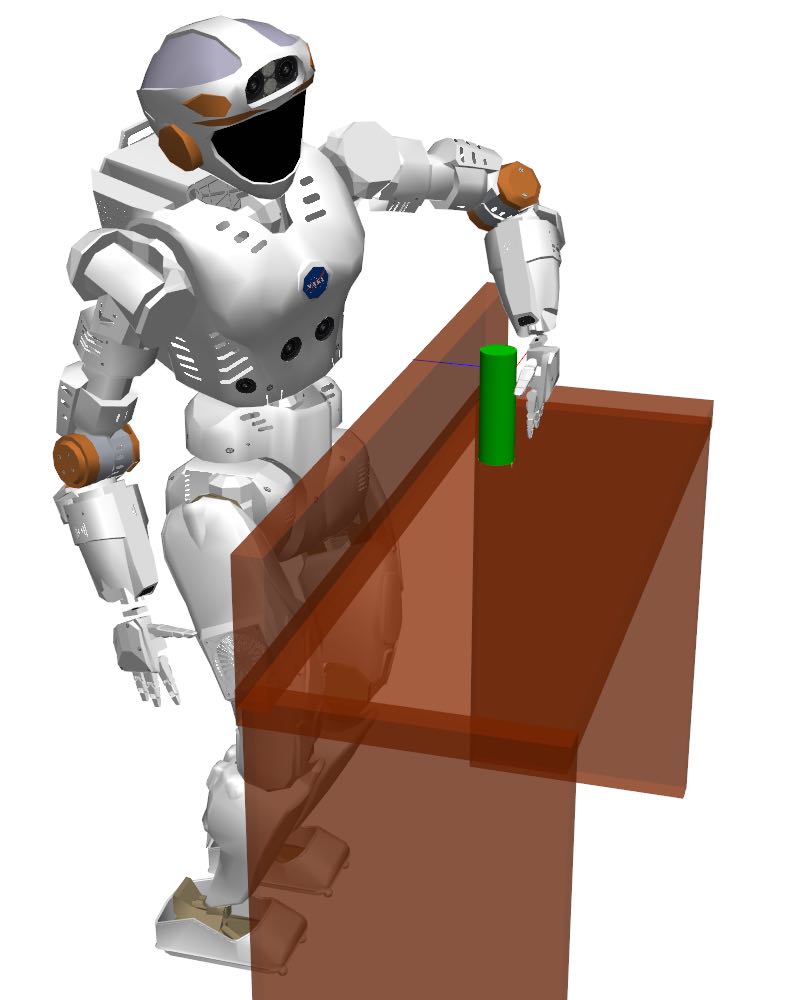}
	\caption{Evaluation tasks, from left to right: task 1, target close to robot; task 2, target far away from robot; and task 3, target behind bar obstacle.}
	\label{fig:tasks}
\end{figure}

\begin{figure}[t]
	\centering
	\subfloat[][Trajectories generated using configuration space sampling.]{
		\includegraphics[width=0.19\textwidth]{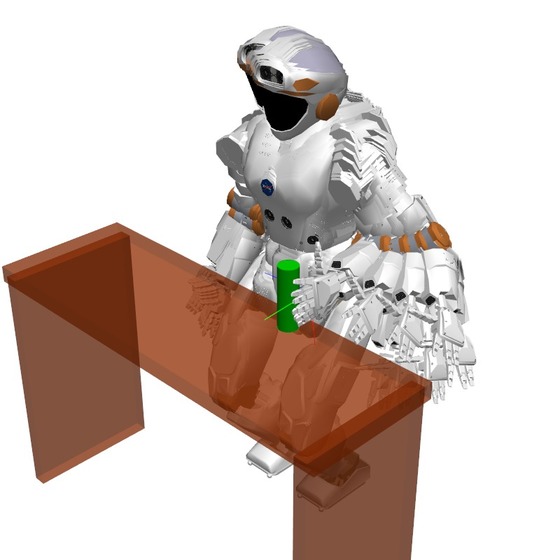}\quad
		\includegraphics[width=0.19\textwidth]{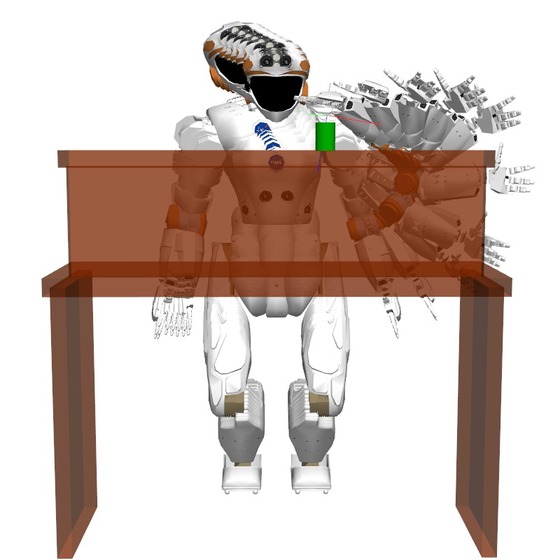}
	}
	 
	\subfloat[][Trajectories generated using end--effector space sampling.]{
		\includegraphics[width=0.19\textwidth]{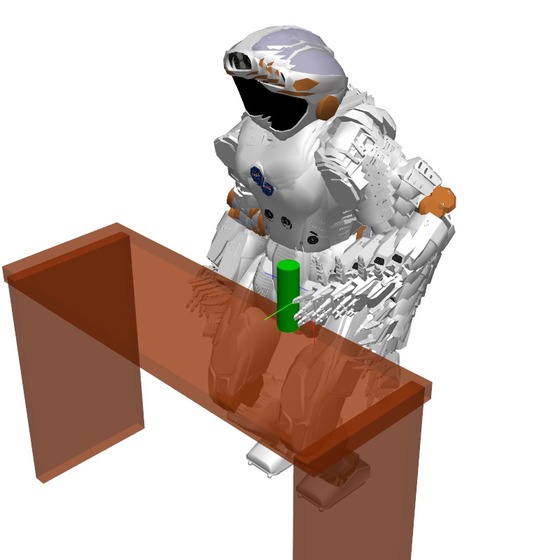}\quad
		\includegraphics[width=0.19\textwidth]{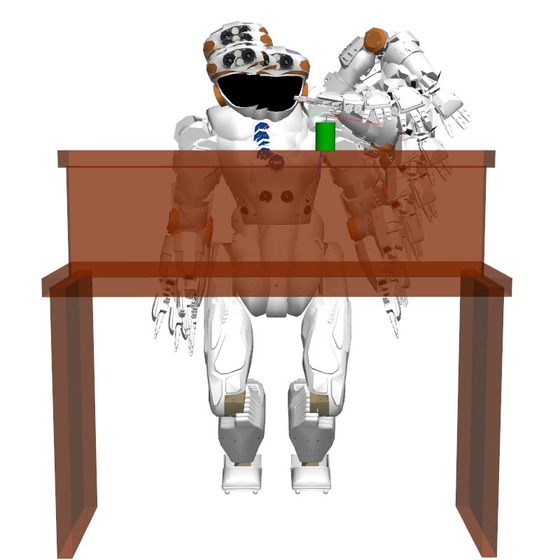}
	}	
	\caption{Whole--body motion plans generated using different sampling spaces. The task is identical for each column. In general, configuration space sampling leads to shorter trajectory length; end--effector space sampling leads to shorter end--effector traverse distance.}
	\label{fig:spacediff}
\end{figure}

\begin{figure*}[t]
	\centering
	\subfloat[Reaching motion on the NASA Valkyrie robot.]{
		\includegraphics[width=0.49\textwidth]{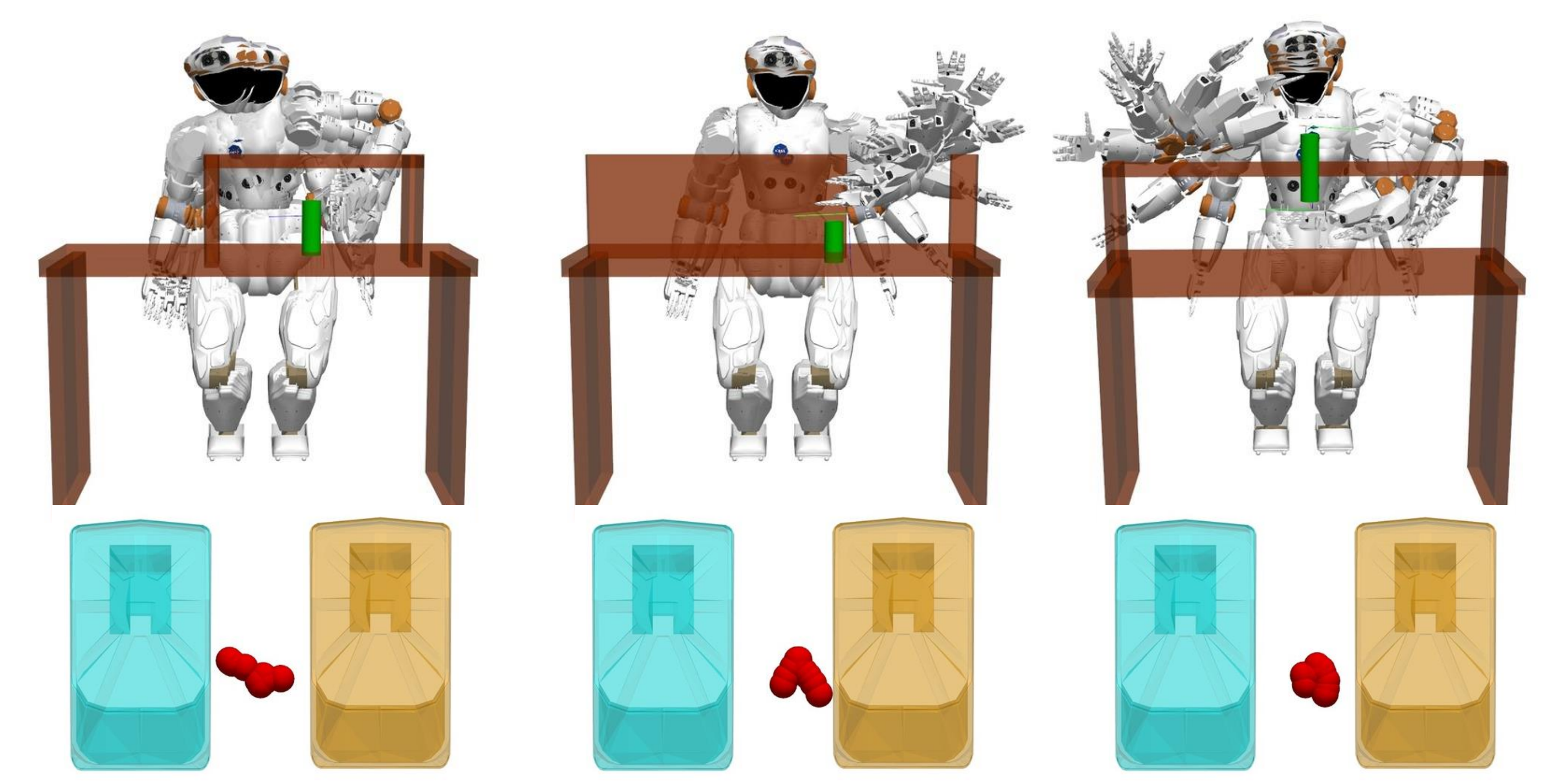}
		\label{fig:val_complex}			
	}
	\subfloat[][Reaching motion on the Boston Dynamics Atlas robot.]{
		\includegraphics[width=0.49\textwidth]{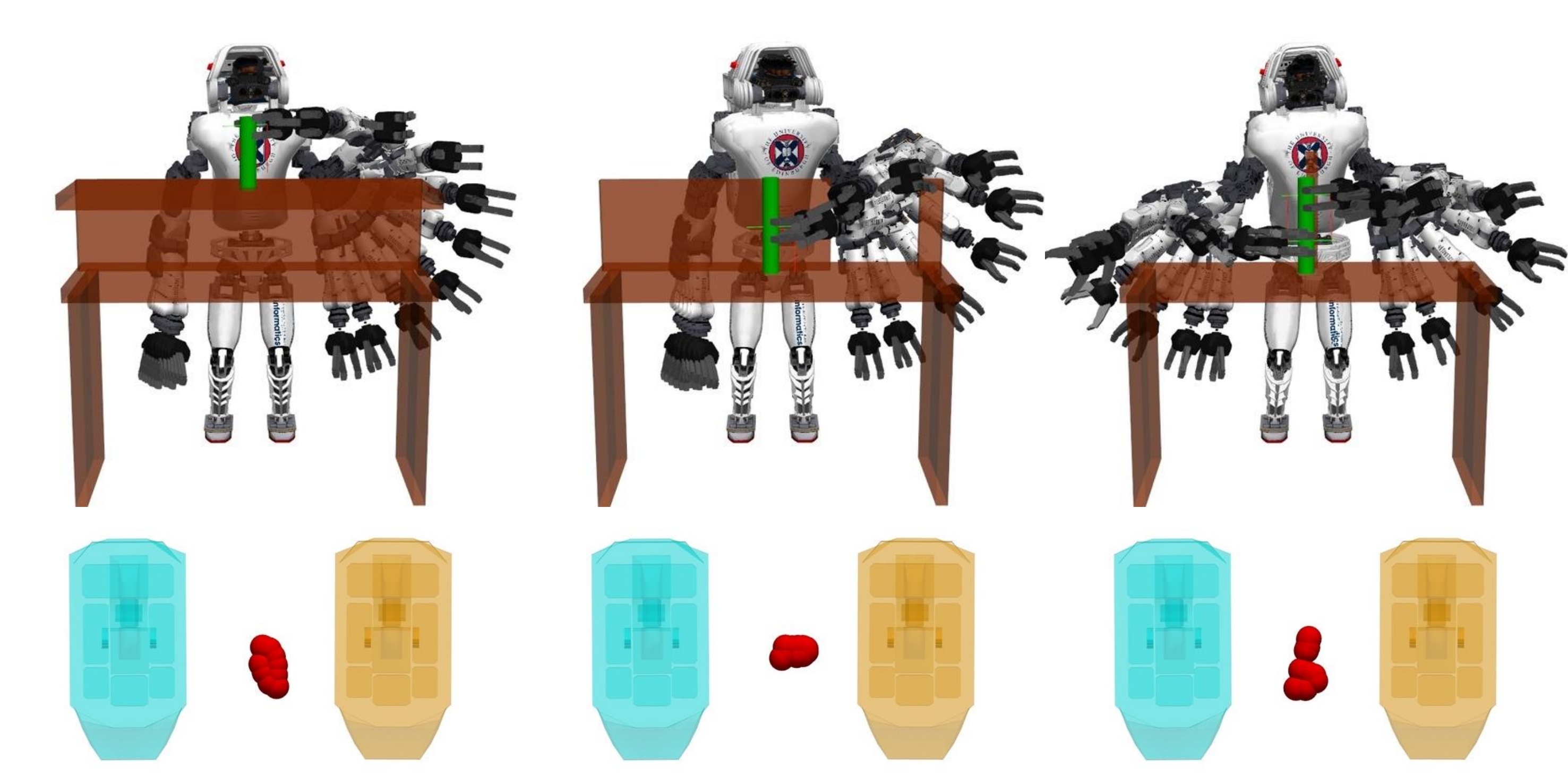}		
		\label{fig:atlas_complex}			
	}
	\caption{Collision--free whole--body motion generated in different scenarios with different robot models. The corresponding CoM trajectories are illustrated in the second row (red dots). The framework is setup so that one can easily switch to new robot platforms without extensive preparing procedures.}
	\label{fig:complex}
\end{figure*}

We setup three different scenarios, from easy to hard, as illustrated in Fig.~\ref{fig:tasks}, to evaluate the performance of different algorithms in different sampling spaces. Unfortunately, the evaluation suggests that standard unidirectional algorithms are unable to solve these problems (within a time limit of $100$ seconds). Without bi--directional search, the high dimensional humanoid configuration space is too complex for sampling--based methods to explore. Table~\ref{table:results} highlights the results using four different bidirectional approaches. Note that when sampling in end--effector space, only RRT--Connect is able to find a valid solution in the given time, other bidirectional search algorithms like BKPIECE and SBL are also unable to find valid trajectories. The result indicates that RRT--Connect sampling in configuration space is the most efficient and the most robust approach for solving humanoid whole--body motion planning problems. It requires the least exploration, thus bypassing expensive online IK queries. Algorithms like BKPIECE and SBL use low--dimensional projections to bias the sampling, however, the default projections which are tuned for mobile robots and robotic arms do not scale up to high DoF humanoid robots, which leads to long planning time and trajectories with high costs. This can be improved by better projection bias, but it is non--trivial to find a suitable bias without fine tuning. Also, the trajectories generated using RRT--Connect are shorter, meaning that the motion is more stable and robust. It is worth mentioning that RRT--Connect takes longer time to plan when sampling in the end--effector space than it does in the configuration space, but the planned trajectories have shorter end--effector and CoM traverse distances. In some scenarios where planning time is not critical, one choose to use RRT--Connect in end--effector space to generate trajectories with shorter end--effector traverse distance. These results also suggest that the whole--body IK computation dominates the planning time. This is in contrast with classical SBP problems where collision--detection is the the most time consuming component. However, the IK solver is necessary for keeping balance, as shown in Fig.~\ref{fig:complex}, where the trajectories' CoM projections are within the support polygon.

In more complex scenarios, such as reaching through narrow passages and bi--manual tasks, most algorithms fail to generate valid trajectories apart from RRT--Connect. As mentioned, some algorithms' performance depends on the biasing methods, e.g. projection bias and sampling bias. However, it is non--trivial to find the appropriate bias for humanoids that would generalize across different tasks. Fig.~\ref{fig:complex} highlights some examples of reaching motion in more complex scenarios with different robot models. As stated earlier, this work focuses on generalising SBP algorithms for humanoids, where as one can easily setup the system on new robot platforms. For instance, one can easily switch from Valkyrie (Fig.~\ref{fig:val_complex}) to Atlas (Fig.~\ref{fig:atlas_complex}) in minutes without extensive pre--processing and setup procedures.

In order to test the reliability and robustness of the proposed method, we applied our work on the Valkyrie robot accomplishing reaching and grasping tasks in different scenarios, as highlighted in Fig~\ref{fig:realrobot}. During practical experiments, the collision environment is sensed by the on–board sensor and represented as an octomap \cite{hornung13auro}. The experiment results show that our method is able to generate collision--free whole--body motion plans that can be executed on full--size humanoid robot to realise practical tasks such as reaching and grasping. A supplementary video of the experiment results can be found at {\href{https://youtu.be/W48miMKWnW4}{\url{https://youtu.be/W48miMKWnW4}}.

\begin{figure*}[t]
	\centering
	\subfloat[][Reach and grasp target on table without facing target.]{
		\includegraphics[width=.99\textwidth]{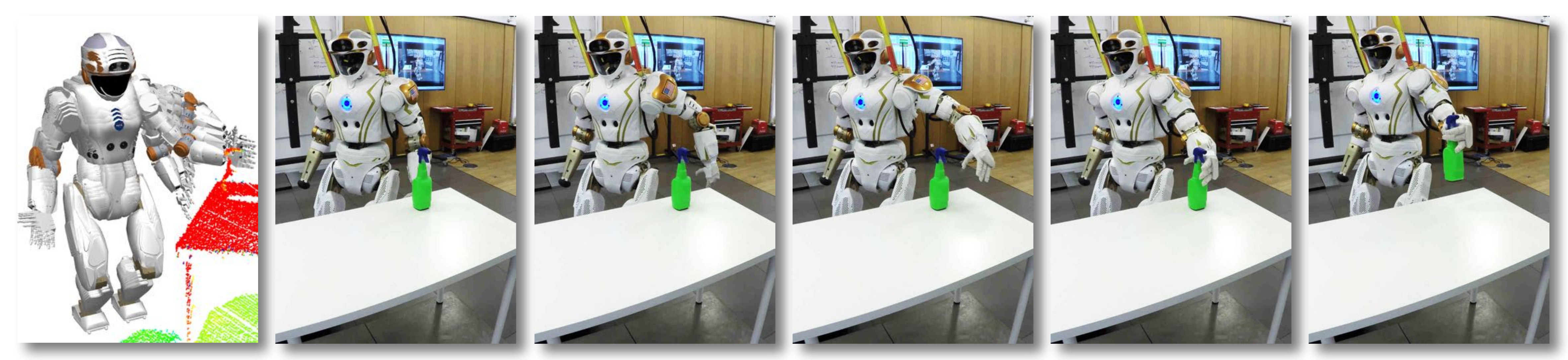}
		\label{fig:showcase1}
	}
	
	\subfloat[][Reach and grasp target on top of box.]{
		\includegraphics[width=.99\textwidth]{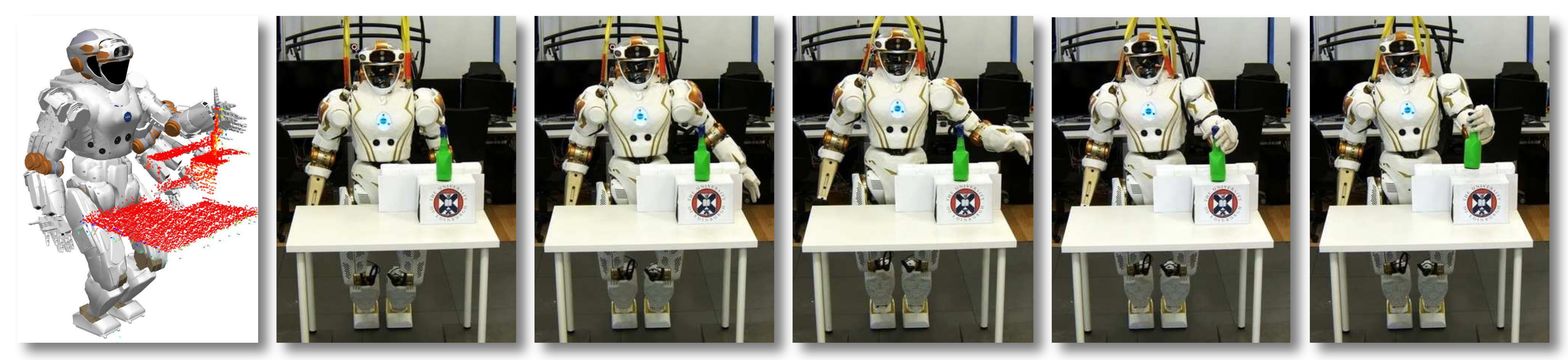}
		\label{fig:showcase2}
	}
	
	\subfloat[][Reach and grasp target on top of shelf.]{
		\includegraphics[width=.99\textwidth]{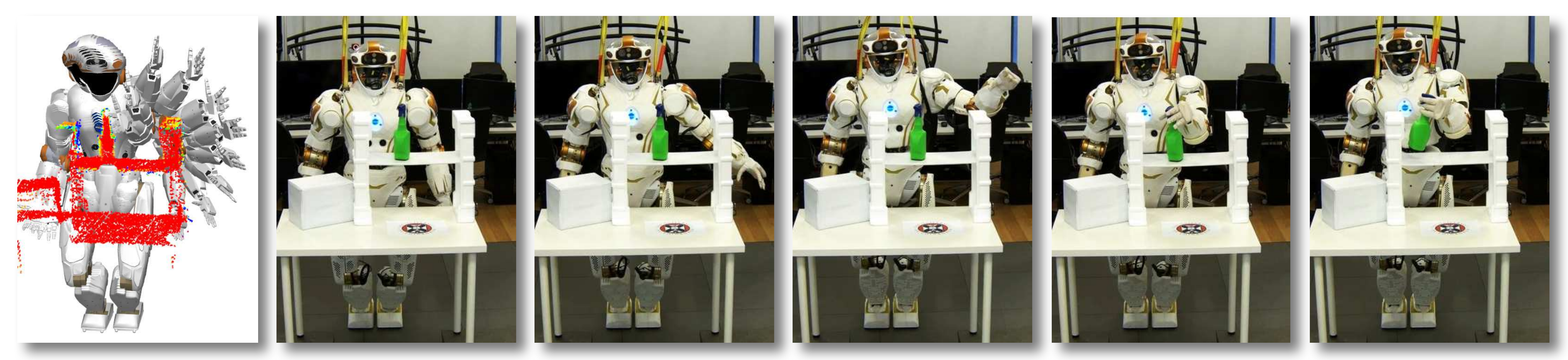}
		\label{fig:showcase3}
	}
	\caption{Collision--free whole--body motion execution on the NASA Valkyrie humanoid robot. In each scenario, the first figure highlights the motion plan, followed by execution snapshots.}
	\label{fig:realrobot}
\end{figure*}

\section{Conclusion}
In this paper we generalise the key components required by sampling--based algorithms for generating collision--free and balanced whole--body trajectories for humanoid robots. We show that by using the proposed methods, standard algorithms can be invoked to directly plan for humanoid robots. We also evaluate the performance of different algorithms on solving planning problems for humanoids, and point out the limitations of some algorithms. A variety of different scenarios are tested showing that the proposed method can generate reliable motion for humanoid robots in different environments. This work can be transferred to different humanoid robot models with easy setup procedure that can be done in very a short period of time, without extensive pre-computation for adapting the existing algorithms to different robot models, as we have tested on the 36 DoF Boston Dynamics Atlas and the 38 DoF NASA Valkyrie robots. In particular, we applied this work on the Valkyrie robot accomplishing different tasks, showing that the proposed method can generate robust collision--free whole--body motion that can be executed on real robots.

The result in Table~\ref{table:results} shows that the whole--body IK solver dominates over 95\% of the online computation time, which currently only runs on a single--thread but can be parallelised on multi--threaded CPU/GPU. The future work will include investigating parallelised implementation of the IK solver on GPU to bootstrap sampling and interpolation. This will make the state space exploration more efficient, so that other standard algorithms may be able to find valid solutions within the same time window.

\bibliographystyle{ieeetr}
\bibliography{bib}
\end{document}